\documentclass[nopreprintline,1p,times]{elsarticle}

\usepackage{amsmath,amssymb}
\usepackage{booktabs}
\usepackage{graphicx}
\graphicspath{{route_C_figures/}{route_B_figures/}{rebuild_results_2026-08-29_v1/figures/}}
\usepackage[colorlinks=true,linkcolor=blue,citecolor=blue,urlcolor=blue,hypertexnames=false]{hyperref}
\usepackage{multirow}
\usepackage{microtype}
\usepackage{placeins}
\usepackage{float}
\usepackage{algpseudocode}
\newfloat{algorithm}{tbp}{loa}
\floatname{algorithm}{Algorithm}

\providecommand{\orcid}[1]{\href{https://orcid.org/#1}{\texttt{#1}}}

\journal{Applied Soft Computing}

\begin{document}

\begin{frontmatter}

\title{Correctness and Predictive Uncertainty in Bayesian TSK Fuzzy Regression: A Reproducible Evaluation on Tabular and Building-Meter Data}

\author[addr]{Zhiren Xiao~\orcid{0009-0008-2164-4557}\corref{cor1}}
\address[addr]{Guangdong University of Finance, Guangzhou, China}
\ead{241734106@m.gduf.edu.cn}
\cortext[cor1]{Corresponding author}

\begin{abstract}
Uncertainty estimates from fuzzy regression are informative only when the implementation is correct and the comparison protocol makes the sources of variation visible.
We revisit a first-order Takagi--Sugeno--Kang (TSK) regression implementation and rebuild its dense, conjugate Bayesian, BIC-based, and rule-level spike-and-slab estimators using corrected code.
The corrected pipeline freezes the Gaussian membership partition at training time, uses a scaled conjugate prior, evaluates firing strengths in the log domain, and reports direct posterior-predictive quantiles for Bayesian methods.
Across 30 outer 80/20 splits on three targets from two UCI regression datasets, corrected TSK-LS obtains $R^2=0.957$, $0.923$, and $0.786$ on Energy-Heating, Energy-Cooling, and Concrete, respectively; rule-level Gibbs sampling with Bayesian model averaging obtains $0.957$, $0.922$, and $0.784$.
We then evaluate the same workflow on a bounded chronological sample of the Building Data Genome 2 meter data that covers 158 buildings across 17 sites.
In a common-unit evaluation, every method is scored on the same 47,400 test rows and every trained model uses 94,800 train--validation rows.
Persistence has RMSE $27.710$, compared with $29.083$ for both TSK-LS and Bayesian TSK; the pre-specified building-level mean RMSE difference between Bayesian TSK and persistence is $+0.0447$ with a 95\% site-cluster bootstrap interval of $[-1.5536,1.5665]$.
The Bayesian TSK 90\% intervals have PICP $0.980$ and MPIW $116.469$.
Taken together, the evidence supports a narrower conclusion: spike-and-slab model averaging is a reproducible implementation option, but it does not provide a demonstrated accuracy or calibration advantage across the tested tabular and building-meter settings.
\end{abstract}

\begin{keyword}
uncertainty quantification \sep prediction intervals \sep Bayesian inference \sep spike-and-slab prior \sep Bayesian model averaging \sep Takagi--Sugeno--Kang fuzzy system \sep building energy
\end{keyword}

\end{frontmatter}

%============================================================
\section{Introduction}
\label{sec:introduction}

Engineering decisions often require more than a point estimate: uncertainty must remain interpretable and reproducible when data are limited.
Rule-based regressions with linear consequents, including Takagi--Sugeno--Kang (TSK) fuzzy systems \cite{takagi1985fuzzy}, offer that combination of nonlinear approximation and a compact rule structure.
Their linear consequents also make Bayesian treatment natural, because posterior distributions, prediction intervals, and model selection can be defined within the same representation. Building-energy simulation illustrates the motivating data regime: each simulation run is computationally expensive, yielding training sets of only hundreds of samples \cite{tsanas2012accurate}.

Measured building meters pose a complementary challenge. Targets are observed over time, buildings differ in use and site, and missing windows or schedule changes can alter the effective prediction problem. The Building Data Genome Project 2 provides an open benchmark with hourly measurements from non-residential buildings across multiple sites and years \cite{miller2020bdg2}; it lets us examine the same uncertainty workflow under chronological extrapolation rather than only independent tabular splits.

The methodological question is deliberately narrower than a general claim about sparse fuzzy regression.
Spike-and-slab priors \cite{mitchell1988bayesian,george1993variable} represent uncertain rule activation, while Bayesian TSK models and prediction-interval methods already provide several probabilistic alternatives \cite{gu2017bayesian,gu2018bayesian,liu2017bayesianzero,miskony2018predictionintervals}.
We therefore ask whether a rule-level Gibbs/BMA implementation changes predictive accuracy or uncertainty once the fuzzy partition, prior parameterization, and numerical linear algebra are made explicit.

The question also exposes an implementation constraint.
The spreads of Gaussian membership functions define the fuzzy partition, so they should be estimated from training data and reused at inference.
Recomputing them from query data can change the design matrix after fitting.
We reproduce this failure mode as an audit, but do not use its unstable output as a corrected-versus-buggy effect size.
All headline comparisons use the frozen-partition implementation and are regenerated from the same code.

We use a block-Gibbs sampler for rule-level spike-and-slab coefficients and predict by Bayesian model averaging (BMA), allowing posterior draws to retain both within-model parameter variation and between-configuration variation.

We compare this estimator with dense TSK least squares, a dense conjugate posterior, the BIC-based thresholded comparator retained for continuity, Gaussian-process and support-vector references, and a split-conformal TSK interval. The evaluation also covers measured building-meter data with a protocol-defined next-hour target, chronological split, and warm-start site holdout.

The contribution is an evidence-led evaluation of this pipeline: correctness gates, proper probabilistic scores, real-building temporal and site behavior, and explicit MCMC and generalization limits are treated as parts of the result.

Our contributions are fourfold:
\begin{enumerate}
\item \textbf{Implementation correctness and traceability.} We make the FCM exponent, frozen membership spreads, no-intercept design, scaled prior, stable covariance calculation, log-domain firing strengths, and direct predictive quantiles explicit, and test them with small-\(R\) analytic and enumeration checks.
\item \textbf{Rebuilt predictive comparison.} We regenerate 30-split results and report point metrics, coverage, interval width, WIS, and CRPS for the TSK, Gaussian-process, and conformal comparators. The rebuilt evidence does not support a claim that Gibbs/BMA dominates the dense or conjugate TSK baselines.
\item \textbf{Real-building evaluation.} We apply the same audited workflow to a bounded BDG2 chronological sample and report next-hour point prediction, posterior-predictive intervals, warm-start site holdout, missing-window eligibility, and runtime boundaries.
\item \textbf{Evidence-bounded sparsity and uncertainty conclusions.} We examine rule-level and coefficient-level sparsity in low-dimensional benchmarks, a controlled irrelevant-feature probe, a real high-dimensional probe, and the BDG2 application setting. The results separate accuracy, coverage, interval width, finite-chain diagnostics, and application scope rather than asserting a universal sparsity or calibration advantage.
\end{enumerate}

Section~\ref{sec:related} reviews related work. Section~\ref{sec:method} presents the TSK model, the membership bug, and the Gibbs-based Bayesian estimators. Section~\ref{sec:setup} describes the controlled benchmarks and the real-building protocol. Section~\ref{sec:results} reports the corrected baseline, the main comparisons, calibration, site behavior, and the sparsity boundary. Section~\ref{sec:discussion} discusses implications and limitations, and Section~\ref{sec:conclusion} concludes.

%============================================================
\section{Related Work}
\label{sec:related}

\subsection{Bayesian TSK Fuzzy Systems}

Bayesian TSK research spans probabilistic classifiers \cite{gu2017bayesian}, joint structure and parameter learning \cite{gu2018bayesian}, zero-order regression models \cite{liu2017bayesianzero}, and Bayesian calibration of computer models \cite{wang2021bayesian}.
Prediction intervals have likewise been constructed for adaptive neuro-fuzzy systems, evolving fuzzy systems, and industrial Takagi--Sugeno soft sensors \cite{miskony2018predictionintervals,cartagena2024evolving,cartagena2023asoc,blazic2024pisensor}.
These studies provide context for the dense conjugate posterior used as a reference here. Our comparison instead isolates the behavior of an explicit rule-level spike-and-slab layer within the same TSK representation.

\subsection{Sparse TSK and Bayesian Variable Selection}

On the frequentist side, sparse TSK systems use embedded feature selection, group regularization, or rule extraction \cite{kbs2024embedded,neuralnet2024sparsetsk,xue2023dgaletsk}.
Recent high-dimensional TSK work also explores multi-head representations \cite{bian2025mhtsk}.
These approaches address different questions from the present rule-level prior: they emphasize point prediction, feature selection, or architecture, whereas we evaluate posterior model averaging for uncertainty.

Spike-and-slab priors and Bayesian model averaging have a broad methodological literature \cite{mitchell1988bayesian,george1993variable,ishwaran2005spike,hoeting1999bayesian}.
Information-criterion and Laplace approximations remain useful comparators, although their reliability depends on the problem \cite{gelfand1994bayesian,chipman2001practical,kasprzak2025laplace}.
Den{\oe}ux's ENNreg \cite{denoeux2023ennreg} provides a belief-function alternative to probabilistic fuzzy regression with prediction intervals.

Distribution-free alternatives such as \emph{split} conformal prediction \cite{vovk2005algorithmic,lei2018distributionfree} provide finite-sample coverage without posterior sampling.
They require a separate calibration split and do not provide the posterior-predictive decomposition used by BMA, so split conformal is included as a complementary rather than interchangeable reference.

\subsection{Real-Building and Temporal Evaluation}

Measured building-energy data add temporal dependence, heterogeneous operating contexts, and missing observation windows to the uncertainty problem.

The full Building Data Genome Project 2 documents hourly meter data from non-residential buildings across 19 sites and two years and was released for prediction benchmarking and related building analytics \cite{miller2020bdg2}. Recent probabilistic electric-load forecasting in ASOC also treats temporal dependence and interval construction as explicit parts of the forecasting problem \cite{zhang2025probabilisticload}.

The setting complements simulated building configurations \cite{tsanas2012accurate}: it supports chronological and site-held-out evaluation, but it does not by itself establish operational deployment or energy savings. We use it to test how the evaluation protocol behaves in a measured setting and to keep the application claims bounded.

\subsection{Reproducibility in Soft Computing}

Silent implementation errors are a recognized source of reproducibility failure in applied machine learning \cite{pineau2021improving}.
In fuzzy regression, a mismatch between the training and inference partitions changes the design matrix after the coefficients have been fitted.
Membership-parameter persistence is therefore treated as an explicit implementation invariant, with its failure mode reported separately from the corrected benchmark.

This perspective connects reproducibility to scientific interpretation. A predictive interval can be numerically well formed while still describing the wrong fitted object if the representation changes between training and inference. Conversely, a stable implementation does not imply that its posterior intervals are calibrated or that a selected rule set improves point prediction. These are different questions and require different evidence.

The present evaluation therefore keeps three distinctions visible throughout the comparison. First, the fuzzy representation is fixed before any model-specific inference is compared. Second, posterior-predictive uncertainty, conformal coverage, and interval width are reported as related but non-interchangeable quantities. Third, benchmark averages, temporal samples, and site-held-out folds are interpreted at their own observational units. This organization turns implementation details into an auditable evaluation object rather than treating them as preprocessing choices that can be omitted from the scientific account.

%============================================================
\section{Methodology}
\label{sec:method}

\subsection{TSK Fuzzy System}
\label{sec:tsk}

For input $\mathbf{x}\in\mathbb{R}^d$ and scalar target $y$, a TSK system with $R$ rules is
\begin{equation}
 y = \sum_{j=1}^{R} w_j(\mathbf{x})\,\big(\beta_{j0}+\beta_{j1}x_1+\cdots+\beta_{jd}x_d\big) + \varepsilon,\quad
 \varepsilon\sim\mathcal{N}(0,\sigma^2),
 \label{eq:tsk}
\end{equation}
where $w_j(\mathbf{x})=\prod_{i=1}^{d}\mu_{A_{ji}}(x_i)\big/\sum_{k=1}^{R}\prod_{i=1}^{d}\mu_{A_{ki}}(x_i)$ is the normalized firing strength of rule $j$, and $A_{ji}$ are Gaussian antecedent fuzzy sets.
\begin{equation}
 \mu_{A_{ji}}(x_i)=\exp\!\Big(-\tfrac{(x_i-c_{ji})^2}{2 s_{ji}^2}\Big),
 \label{eq:membership}
\end{equation}
Centers $c_{ji}$ are obtained from fuzzy $c$-means clustering \cite{bezdek1984fcm}; spreads $s_{ji}$ are the standard deviations of the training points assigned to each cluster, scaled by $s$. The implementation adds a fixed $0.01$ offset after scaling to avoid zero spreads; the Route C implementation also enforces a $0.01$ lower bound. With the weighted design matrix $\boldsymbol{\Phi}\in\mathbb{R}^{n\times R(d+1)}$ and the stacked consequent vector $\boldsymbol{\beta}\in\mathbb{R}^{R(d+1)}$, Eq.~\eqref{eq:tsk} becomes the linear model $\mathbf{y}=\boldsymbol{\Phi}\boldsymbol{\beta}+\boldsymbol{\varepsilon}$.

\subsection{The Membership-Spread Bug}
\label{sec:bug}

The spreads $s_{ji}$ in Eq.~\eqref{eq:membership} define the fuzzy partition, so training and inference must use the same values.
The audited defect recomputes them at prediction by passing query inputs and placeholder cluster labels to the membership routine.
This changes the partition after $\boldsymbol{\beta}$ has been fitted and can make predictions numerically unstable.
The defective path is therefore treated as a failure-mode reproduction rather than a valid benchmark, and its output is not used as a headline effect size.
The corrected procedure stores the training-time centers and spreads and reuses them verbatim at inference, as in Algorithm~\ref{alg:tsk}.

\begin{algorithm}[tbhp]
\caption{Corrected TSK training and inference}
\label{alg:tsk}
\begin{algorithmic}[1]
\Require Training data $\{\mathbf{x}_i,y_i\}_{i=1}^n$, rule count $R$
\State $(\mathbf{c},\mathbf{u}) \gets \mathrm{FCM}(\mathbf{X}, R)$ \Comment{fuzzy $c$-means}
\State $\mathbf{s} \gets \{\text{per-cluster standard deviations from training points, scaled by } s\}$
\State Store $(\mathbf{c},\mathbf{s})$ \Comment{\emph{fix the partition at training time}}
\State Build $\boldsymbol{\Phi}$ via Eqs.~\eqref{eq:tsk}--\eqref{eq:membership} using $(\mathbf{c},\mathbf{s})$
\State Fit $\boldsymbol{\beta}$ (least squares, conjugate Bayes, or Gibbs; Sections~\ref{sec:conj}--\ref{sec:gibbs})
\State \Return model with frozen $(\mathbf{c},\mathbf{s},\boldsymbol{\beta})$
\end{algorithmic}
\end{algorithm}

\subsection{Evaluation Object and Invariants}
\label{sec:invariants}

The fitted object in this study has four linked layers: an antecedent partition, a weighted linear design, a parameter distribution, and a predictive distribution. The antecedent layer supplies the centers and spreads that determine each rule's firing strength. The weighted design then maps those firing strengths to the linear consequents.

The conjugate or Gibbs layer describes uncertainty in the consequent coefficients and noise variance, and the predictive layer propagates those draws to a new input. A comparison is meaningful only when these layers refer to the same stored partition and the same target scale.

This dependency explains why the membership-spread defect is more than a numerical nuisance. Re-estimating spreads from query inputs changes the antecedent layer and therefore changes the design row used by a coefficient vector that was fitted under a different representation. The resulting output cannot be interpreted as a less accurate version of the same model without first establishing representation equivalence.

The corrected implementation consequently treats centers and spreads as part of the fitted model state, just as the consequent coefficients and posterior scale are part of that state.

The same logic determines how the uncertainty comparisons are reported. A posterior-predictive interval combines parameter and noise variation under a specified model, whereas a split-conformal interval is calibrated from held-out residuals under a separate protocol.

Coverage, width, proper scores, and point error can therefore move in different directions without contradiction. The analysis preserves those distinctions instead of reducing them to one composite ranking. The analytic forced-active check, synthetic enumeration, finite-chain diagnostics, and cross-target comparisons each test a different layer of this evaluation object; none is presented as evidence for a new estimator beyond the specified TSK pipeline.

\subsection{Conjugate Bayesian Baseline}
\label{sec:conj}

The dense Bayesian TSK places the scaled conjugate Gaussian prior $\boldsymbol{\beta}\mid\sigma^2\sim\mathcal{N}(\mathbf{0},\sigma^2\tau^2\mathbf{I})$ and an inverse-gamma prior $\sigma^2\sim\mathrm{InvGamma}(0.01,0.01)$ on the noise variance. With $\mathbf{P}=\boldsymbol{\Phi}^{\top}\boldsymbol{\Phi}+\tau^{-2}\mathbf{I}$, the posterior is available in closed form:
\begin{equation}
 \boldsymbol{\beta}\mid\mathcal{D},\sigma^2 \sim \mathcal{N}\big(\mathbf{P}^{-1}\boldsymbol{\Phi}^{\top}\mathbf{y},\ \sigma^2\mathbf{P}^{-1}\big),
\end{equation}
and the predictive variance conditional on $\sigma^2$ at a new design row $\boldsymbol{\phi}^*$ is $\sigma^2 + \boldsymbol{\phi}^*(\sigma^2\mathbf{P}^{-1})\boldsymbol{\phi}^{*\top}$.
After integrating $\sigma^2$, the implementation reports the corresponding Student-$t$ posterior-predictive quantiles.

This closed-form model is a dense reference and a correctness target for the Gibbs updates when all rules are forced active. The forced-active check is complemented by a small-$R$ enumeration check of the rule-inclusion update; both are reported in the Supplementary Material.

The weighted TSK design can be ill-conditioned, so the implementation uses symmetric solves, explicit diagonal stabilization, and no additional regression intercept after the rule-weighted intercept columns have been constructed. The full condition-number and numerical-stability descriptors are retained in the phase-diagram output rather than summarized by a single benchmark-wide condition number.

\subsection{Spike-and-Slab Prior and Gibbs Sampling}
\label{sec:gibbs}

We decompose $\boldsymbol{\beta}$ into $R$ blocks $\boldsymbol{\beta}_j\in\mathbb{R}^{d+1}$ and place a spike-and-slab prior on each block:
\begin{equation}
 \boldsymbol{\beta}_j \mid \gamma_j,\sigma^2,\tau^2 \sim
 \begin{cases}
  \mathcal{N}(\mathbf{0},\sigma^2\tau^2\mathbf{I}_{d+1}), & \gamma_j=1 \ \ \text{(slab)},\\[3pt]
  \delta_{\mathbf{0}}, & \gamma_j=0 \ \ \text{(spike)},
 \end{cases}
 \label{eq:spikeslab}
\end{equation}
with $\gamma_j\sim\mathrm{Bernoulli}(\pi)$. We update $\boldsymbol{\gamma}$ with a block-Gibbs sampler rather than with the BIC criterion used by the comparator.
The sampler targets the specified posterior conditional on a finite Monte Carlo run. Its estimates therefore retain Monte Carlo error, which is quantified with the convergence diagnostics of Section~\ref{sec:setup}.

For block $j$, let $\mathbf{r}_{-j}=\mathbf{y}-\sum_{k\neq j}\boldsymbol{\Phi}_k\boldsymbol{\beta}_k$ be the residual excluding rule $j$, and let $\mathbf{A}_j=\boldsymbol{\Phi}_j^{\top}\boldsymbol{\Phi}_j$.
The marginal likelihood of $\mathbf{r}_{-j}$ under $\gamma_j=1$ (integrating out $\boldsymbol{\beta}_j$) is Gaussian with covariance $\sigma^2(\mathbf{I}+\tau^2\boldsymbol{\Phi}_j\boldsymbol{\Phi}_j^{\top})$.
The determinant and quadratic form are evaluated in $\mathcal{O}(n(d+1)^2)$ using the matrix-determinant and Woodbury identities. The full conditional for $\gamma_j$ is
\begin{equation}
 P(\gamma_j=1\mid\cdot)=\sigma\!\Big(\log\tfrac{\pi}{1-\pi}
  -\tfrac12\log\det(\mathbf{I}+\tau^2\mathbf{A}_j)
  +\tfrac{\tau^2}{2\sigma^2}\mathbf{z}_j^{\top}(\mathbf{I}+\tau^2\mathbf{A}_j)^{-1}\mathbf{z}_j\Big),
 \label{eq:gamma}
\end{equation}
where $\mathbf{z}_j=\boldsymbol{\Phi}_j^{\top}\mathbf{r}_{-j}$ and $\sigma(\cdot)$ is the logistic function.

Conditional on $\gamma_j=1$, we draw $\boldsymbol{\beta}_j\sim\mathcal{N}\big((\mathbf{A}_j+\tau^{-2}\mathbf{I})^{-1}\mathbf{z}_j,\ \sigma^2(\mathbf{A}_j+\tau^{-2}\mathbf{I})^{-1}\big)$; otherwise $\boldsymbol{\beta}_j=\mathbf{0}$.
The noise variance $\sigma^2$ is then drawn from its inverse-gamma full conditional with the corresponding active-block prior quadratic form.
A coefficient-level variant applies the same prior independently to each scalar coefficient $\beta_i$ (stochastic search variable selection, SSVS \cite{george1993variable}).
For numerical stability, sampled precision matrices receive a small diagonal jitter; coefficients are initialized from a ridge-regularized least-squares fit with no additional intercept, and the sampled noise variance is floored at $10^{-4}$.

\subsection{Bayesian Model Averaging for Prediction}
\label{sec:bma}

Prediction uses BMA rather than conditioning on a single selected model \cite{hoeting1999bayesian}. For BMA practice and its pitfalls, see \cite{fragoso2018bayesian}. Given $T$ posterior draws $\{\boldsymbol{\beta}^{(t)},\sigma^{2(t)}\}_{t=1}^{T}$, the predictive mean is the mixture mean $\hat{y}^*=\frac1T\sum_t \boldsymbol{\phi}^*\boldsymbol{\beta}^{(t)}$, and the predictive variance follows from the law of total variance:
\begin{equation}
 \mathrm{Var}(y^*\mid\mathbf{x}^*,\mathcal{D})
 = \underbrace{\tfrac1T\sum_t \sigma^{2(t)}}_{\text{aleatoric}}
 + \underbrace{\mathrm{Var}_t\big(\boldsymbol{\phi}^*\boldsymbol{\beta}^{(t)}\big)}_{\text{model uncertainty}},
 \label{eq:bma}
\end{equation}

The reported bounds are the direct 2.5th and 97.5th quantiles of this posterior-predictive mixture, not moment-matched Gaussian bounds.
Across posterior draws, the second term captures both between-model (rule-configuration) variation and within-model parameter variation.
The construction therefore retains model-selection uncertainty; whether that uncertainty improves calibration remains an empirical question.

\subsection{Route C Common-Unit Real-Building Protocol}
\label{sec:routec}

For the measured-data evaluation, next-hour cleaned electricity is the target. Predictors comprise three lag values, two past-window statistics, two cyclical hour features, and a weekend indicator. Weather is excluded from the primary model so that the main comparison is not confounded by site-local timestamp alignment. The cohort is selected from BDG2 metadata using target-coverage eligibility and a fixed hash order; neither target quality nor test performance enters selection. Training uses 2016, validation uses January--June 2017, and testing uses July--December 2017. Each building--split contributes 300 evenly spaced valid supervised rows in the selected cohort.

The Route C primary comparison fixes 47,400 common test rows and fits all trained point models on the concatenated training and validation rows, 94,800 observations in total. Ridge hyperparameter selection remains confined to the frozen validation split. The conformal reference uses a separate validation calibration object; this is reported as a calibration-protocol distinction, not as a violation of the final point-model exposure rule. The external site analysis is retained as a complementary warm-start site holdout rather than a cold-start transfer experiment. Missing target or feature windows are excluded without imputation.

%============================================================
\section{Experimental Setup}
\label{sec:setup}

\subsection{Controlled Benchmarks}

The controlled study uses two UCI regression benchmarks and three targets: the Energy Efficiency dataset \cite{tsanas2012accurate} (768 simulated building configurations, 8 geometric features, with heating and cooling loads as two targets) and the Concrete Compressive Strength dataset \cite{yeh1998modeling} (1030 laboratory samples, 8 mix-design features). Features are standardized to zero mean and unit variance. All TSK methods use $R=5$ rules from fuzzy $c$-means ($m=2$), with Gaussian membership spreads equal to the per-cluster standard deviation of training points and scaled by $s=1.5$.
A rule-count sweep ($R=3$--$10$, Supplementary) shows dense accuracy is nearly unchanged from $R=4$ upward on the Energy targets and improves only marginally beyond $R\approx7$ on Concrete, supporting the default $R=5$.

\subsection{Measured Building Benchmark}

The measured-building evaluation uses the cleaned electricity panel from BDG2 v1.0 \cite{miller2020bdg2,miller2020bdg2zenodo}. The target is the next-hour electricity sum in the units supplied by the dataset. The protocol-defined cohort contains 158 eligible buildings across 17 sites, with a minimum target-coverage requirement applied independently to each temporal split. The primary predictors are the eight causal lag, rolling, cyclical-time, and weekend features specified in the Route C configuration. Weather is excluded because the data are recorded in local time and weather alignment requires a separate time-zone rule.

The chronological evaluation trains on 2016, uses January--June 2017 for validation, and tests on July--December 2017. The primary Route C comparison uses 300 evenly spaced valid supervised rows for each of 158 buildings, giving 47,400 common test rows. All trained models use the 47,400 training rows plus the 47,400 validation rows for their final fit. The site analysis uses five fixed round-robin folds and retains explicitly available historical lag values; it is therefore reported as a complementary warm-start site holdout. No raw BDG2 archive is redistributed with the working manuscript.

\begin{table}[htbp]
\centering
\caption{Frozen Route C protocol used for the BDG2 common-unit evaluation.}
\label{tab:bdg2-protocol}
\small
\begin{tabular}{l p{0.62\linewidth}}
\toprule
Item & Frozen choice \\
\midrule
Dataset & BDG2 v1.0; cleaned electricity meter \\
Cohort & Existing outcome-independent rule; 158 buildings across 17 sites \\
Target & Next-hour electricity; unit as supplied by the dataset \\
Features & 8 causal lag, rolling and calendar features; primary weather excluded \\
Temporal split & Train 2016; validation 2017-01--2017-06; test 2017-07--2017-12 \\
Sampling & 300 evenly spaced valid rows per building and split; 47,400 common test rows \\
Final fit & Train+validation for all trained models; 94,800 rows \\
Primary estimand & Equal-building-weight RMSE difference: Bayesian TSK minus persistence \\
Uncertainty & 90\% and 95\% posterior-predictive intervals where available; conformal calibration kept separate \\
\bottomrule
\end{tabular}
\end{table}

Four TSK methods are compared in the controlled benchmark experiments: \textbf{TSK-LS} (dense least squares), \textbf{Bayesian-TSK} (dense conjugate posterior, Section~\ref{sec:conj}), \textbf{TSK-SpikeSlab-BIC} (the BIC-plus-Laplace comparator with hard thresholding), and \textbf{TSK-SpikeSlab-Gibbs} (rule-level Gibbs plus BMA, Sections~\ref{sec:gibbs}--\ref{sec:bma}).
A coefficient-level variant, \textbf{TSK-SSVS}, is used in the sparsity analysis (Section~\ref{sec:sparse}). Random forest, support vector regression, and Gaussian-process regression serve as machine-learning references; the Gaussian process provides a probabilistic interval baseline.

For BDG2, the Route C primary comparison retains persistence, seasonal-naive, Ridge, HistGradientBoosting, TSK-LS, Bayesian TSK, three seed-specific Gibbs TSK runs, and a Ridge conformal reference. Reference implementations use the configurations recorded in the frozen Route C experiment file. Every method is evaluated on the same 47,400 sampled test rows. The trained models use the same 94,800 train-plus-validation rows; the conformal reference keeps its validation calibration object separate. The three Gibbs rows are seed-specific full-exposure runs and are not pooled into a cross-seed posterior.

The random-forest reference uses 300 trees and maximum depth 10. Support vector regression uses an RBF kernel ($C=1$, $\gamma=$ \texttt{scale}), and the Gaussian-process reference uses an RBF-plus-white-noise kernel with hyperparameters optimized by marginal likelihood (\texttt{normalize\_y=True}).

Fuzzy-specific interval-construction methods such as ENNreg \cite{denoeux2023ennreg} and interval-ANFIS \cite{miskony2018predictionintervals} build intervals through evidential combination or interval-network training rather than posterior model selection.
They are complementary to the present comparison.

The rule-level Gibbs sampler uses $1{,}000$ burn-in and $2{,}000$ retained draws in the controlled UCI experiments; the coefficient-level SSVS uses $800$ burn-in and $800$ retained draws. Route C uses 200 burn-in iterations and 500 retained draws for each of three full-exposure seed-specific Gibbs runs.
The Route C primary point estimand is the equal-building-weight mean RMSE difference between Bayesian TSK and persistence. The stored Route B comparison had unequal training and evaluation exposure; Route C resolves that discrepancy for the promoted common-unit result.

The diagnostics show raw coefficient \(\hat R_{\max}=1.061\) for Energy-Cooling and \(1.030\) for Concrete, while the maximum predictive-quantity \(\hat R\) is \(1.004\) and \(1.027\), respectively; the mean ESS for \(\sigma^2\) is \(1{,}639.5\) and \(1{,}740.3\) (Supplementary Table~10).
These diagnostics do not establish perfect mixing, and no formal convergence claim is made for coefficient-level SSVS.
The UCI results are averaged over 30 independent random 80/20 train--test splits with a fixed seed (SEED=42). The Route C BDG2 main evaluation uses one frozen chronological sampled split with a pre-specified building-level paired estimand; the complementary site-held-out table retains the warm-start boundary and is not pooled with the primary estimate.

Bayesian intervals use direct 2.5th and 97.5th posterior-predictive quantiles; conformal intervals use a calibration split within each outer training set.
All spike-and-slab priors use inclusion probability $\pi=0.5$; the rule-level Gibbs and conjugate baselines use slab variance $\tau^2=10^3$, and the coefficient-level SSVS uses $\tau^2=1.0$.
Metrics are root-mean-square error (RMSE), mean absolute error (MAE), $R^2$, prediction-interval coverage probability (PICP), mean prediction-interval width (MPIW), weighted interval score (WIS), and continuous ranked probability score (CRPS) when a predictive distribution is available. BDG2 results additionally report conditional site behavior, active rules, runtime, memory and failure cases.

\subsection{Comparison Logic}
\label{sec:comparisonlogic}

The comparison is organized around the unit at which each protocol makes a claim. For the UCI benchmarks, the unit is a target-specific outer split and the reported aggregate is the mean over 30 fixed 80/20 repetitions. For BDG2 Route C, the primary unit is the building-level test error within the common chronological test rows, with site-cluster bootstrap uncertainty; the complementary site analysis is a fold-level warm-start comparison retained separately. These units are not pooled, because averaging across simulated tabular targets and measured temporal windows would conceal the change in data-generating context that the application evaluation is intended to expose.

Baseline choice follows the same separation. Dense TSK-LS and dense conjugate Bayes test whether rule-level selection changes the fitted representation or predictive uncertainty relative to a non-selected TSK reference. The BIC comparator tests a different selection mechanism under the same broad representation. Gaussian-process and support-vector references indicate whether any observed behavior is specific to the fuzzy model family, while persistence, seasonal-naive, Ridge, and HistGradientBoosting establish application-level references for the measured-building task. The conformal baseline is included to compare a calibrated residual construction with posterior-predictive intervals, not to imply that the two intervals estimate the same object.

Every result is consequently read with its protocol attached. The main UCI table answers whether rule-level Gibbs/BMA changes average predictive behavior under repeated tabular splits. The Route C BDG2 table answers whether the same workflow changes point error or interval behavior when all methods share a common test unit and final training exposure. The site table answers how the complementary measured-data result varies across held-out sites when historical lags remain available. The ablation tables then ask whether prior scale, noise features, rule count, and selection mode change that interpretation. This ordering makes negative and heterogeneous results part of the evidence ladder rather than exceptions to a preferred ranking.

%============================================================
\section{Results}
\label{sec:results}

\subsection{Implementation Audit}
\label{sec:repro}

The implementation audit establishes that membership spreads must persist from training to inference.
Reproducing the defective query-time recomputation path produced numerically unstable predictions, including an extreme negative $R^2$ on Energy-Cooling in the recorded defect run.
We retain that output as a failure-mode diagnostic rather than a corrected-versus-buggy performance estimate.
The comparison below uses only the frozen-partition implementation, so every reported value belongs to one stable protocol.

\subsection{Main Comparison}
\label{sec:main}

Table~\ref{tab:main} presents the primary results generated from the rebuilt JSON output.
Corrected TSK-LS gives $R^2=0.957$, $0.923$, and $0.786$ on Energy-Heating, Energy-Cooling, and Concrete.
Bayesian-TSK gives $0.957$, $0.923$, and $0.786$, while TSK-SpikeSlab-Gibbs gives $0.957$, $0.922$, and $0.784$.
Rule-level sampling with BMA is therefore close to the dense and conjugate TSK references in this protocol, but it does not demonstrate a predictive-accuracy gain.
The BIC comparator likewise remains close to the dense reference ($R^2=0.956$, $0.922$, and $0.786$); the negative-$R^2$ collapse from the earlier implementation is not reproduced after correction.

The non-fuzzy references are stronger on point prediction for these data: the Gaussian process reaches $R^2=0.998$, $0.979$, and $0.889$, while random forest reaches $0.998$, $0.969$, and $0.903$.

\begin{table}[htbp]
\centering
\caption{Main comparison. Values are means over the 30 outer 80/20 splits. RMSE is on the original target scale; PICP is the empirical coverage of the 95\% posterior-predictive interval when available.}
\label{tab:main}
\scriptsize
\setlength{\tabcolsep}{2.5pt}
\resizebox{\textwidth}{!}{%
\begin{tabular}{lccc|ccc|ccc}
\toprule
& \multicolumn{3}{c|}{Energy-Heating} & \multicolumn{3}{c|}{Energy-Cooling} & \multicolumn{3}{c}{Concrete} \\
Method & $R^2$ & RMSE & PICP & $R^2$ & RMSE & PICP & $R^2$ & RMSE & PICP \\
\midrule
TSK-LS & 0.957 & 2.025 & -- & 0.923 & 2.599 & -- & 0.786 & 7.694 & -- \\
Bayesian-TSK & 0.957 & 2.030 & 0.940 & 0.923 & 2.602 & 0.932 & 0.786 & 7.696 & 0.938 \\
TSK-SpikeSlab-BIC & 0.956 & 2.044 & 0.940 & 0.922 & 2.614 & 0.934 & 0.786 & 7.695 & 0.938 \\
TSK-SpikeSlab-Gibbs & 0.957 & 2.033 & 0.940 & 0.922 & 2.607 & 0.934 & 0.784 & 7.732 & 0.938 \\
Random Forest & 0.998 & 0.488 & -- & 0.969 & 1.663 & -- & 0.903 & 5.180 & -- \\
SVR & 0.925 & 2.729 & -- & 0.892 & 3.093 & -- & 0.643 & 9.967 & -- \\
Gaussian Process & 0.998 & 0.470 & 0.931 & 0.979 & 1.353 & 0.938 & 0.889 & 5.535 & 0.939 \\
\bottomrule
\end{tabular}%
}
\end{table}

The cross-target pattern is more informative than any single $R^2$ value. Gibbs/BMA remains close to the dense TSK references on all three targets, but the small differences do not establish a repeatable predictive gain. The non-fuzzy references are stronger on the same point-prediction metrics, which prevents the fuzzy uncertainty representation from being interpreted as an accuracy advantage by default. At the same time, the corrected BIC comparator no longer reproduces the extreme failure behavior of the earlier implementation.

The result is therefore a two-part finding: implementation correctness changes the scientific conclusion about the comparator, while model averaging itself does not produce a demonstrated point-prediction improvement under this protocol.

This pattern also clarifies what the benchmark can and cannot distinguish. The dense and selected TSK models share the frozen antecedent representation, so their close performance is evidence about the incremental value of rule-level selection within that representation. It is not evidence that all sparse fuzzy models behave similarly, nor that the selected posterior is unnecessary in other regimes. The controlled comparison supports a bounded statement about the tested prior, rule count, split protocol, and finite Monte Carlo run; the broader question of when sparse rule structures improve prediction remains open.

\subsection{Measured-Building Evaluation}
\label{sec:bdg2results}

The measured-building benchmark asks whether the controlled-tabular conclusions persist when observations are ordered in time and drawn from heterogeneous sites. Table~\ref{tab:bdg2-main} reports the Route C common-unit result. All methods use the same 47,400 test rows, and every trained model uses 94,800 train-plus-validation rows. Persistence has the lowest point RMSE at $27.710$, compared with $29.083$ for TSK-LS and $29.083$ for Bayesian TSK.

The pre-specified equal-building-weight mean difference between Bayesian TSK and persistence is $+0.044720$ RMSE units. The 95\% site-cluster bootstrap interval is $[-1.553640,1.566517]$, and Bayesian TSK has lower RMSE on 46.8\% of buildings. Under this common-unit protocol, no point-prediction advantage is demonstrated.

Interval behavior separates the point-error comparison. Bayesian TSK gives PICP $=0.979937$ and MPIW $=116.469477$ for the 90\% interval, whereas the Ridge conformal reference gives PICP $=0.900865$ and MPIW $=30.634671$. The posterior interval is wider and overcovers on the common sampled test split; the conformal interval is closer to its nominal level under its own calibration protocol. Because the intervals are different probability objects, coverage and width are reported together rather than collapsed into one ranking.

\begin{table}[t]
\centering
\caption{BDG2 Route C common-unit comparison. All methods use the same 47,400 sampled test rows; trained models use the same 94,800 train-plus-validation rows.}
\label{tab:bdg2-main}
\small
\begin{tabular}{lrrrrr}
\toprule
Method & RMSE $\downarrow$ & MAE $\downarrow$ & $R^2$ $\uparrow$ & PICP$_{90}$ & MPIW$_{90}$ \\
\midrule
Persistence & 27.710 & 7.321 & 0.995 & -- & -- \\
Seasonal naive (24 h) & 86.360 & 16.654 & 0.947 & -- & -- \\
Seasonal naive (168 h) & 110.895 & 18.165 & 0.913 & -- & -- \\
Ridge & 30.463 & 9.187 & 0.993 & -- & -- \\
HistGradientBoosting & 52.042 & 10.480 & 0.981 & -- & -- \\
TSK-LS & 29.083 & 8.676 & 0.994 & -- & -- \\
Bayesian TSK & 29.083 & 8.675 & 0.994 & 0.980 & 116.469 \\
Gibbs TSK (seeds 42--44) & 29.519 $\pm$ 0.379 & 9.430 & 0.994 & 0.980 & 118.140 \\
Ridge conformal & 30.458 & 9.187 & 0.993 & 0.901 & 30.635 \\
\bottomrule
\multicolumn{6}{p{0.96\linewidth}}{\footnotesize Note: Point and interval metrics are computed on the same 47,400 test rows. The Gibbs row is the mean across three seed-specific full-exposure runs, with sample RMSE standard deviation shown; it is not a pooled cross-seed posterior. Bayesian/Gibbs intervals are posterior-predictive; the conformal interval uses its separate validation calibration protocol. A dash denotes an unavailable predictive interval.}
\end{tabular}
\end{table}

\begin{figure}[t]
\centering
\includegraphics[width=\linewidth]{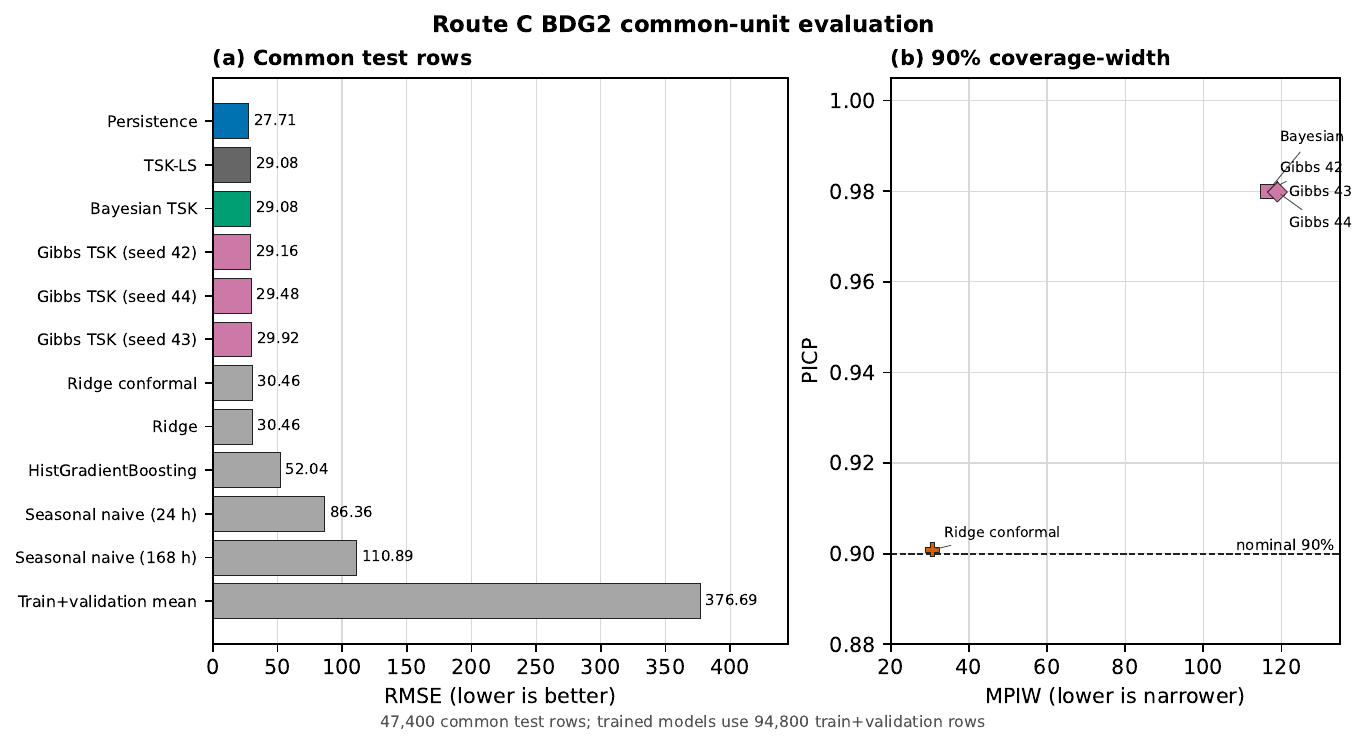}
\caption{Route C common-unit evidence on the BDG2 sampled test split. Panel (a) compares point RMSE on the same 47,400 test rows; panel (b) shows the 90\% coverage--width trade-off for methods with predictive or conformal intervals. Trained models use 94,800 train-plus-validation rows. The result does not represent operational deployment or cold-start transfer.}
\label{fig:bdg2routec}
\end{figure}

Figure~\ref{fig:bdg2routec} summarizes the same comparison. Its point-error panel makes the persistence result visible, while its second panel shows the width cost associated with higher coverage for the posterior and conformal references.

The complementary site-held-out analysis in Table~\ref{tab:bdg2-site} reveals substantial heterogeneity. It is retained from the stored warm-start site protocol rather than treated as part of the Route C primary estimand. Bayesian TSK RMSE ranges from $9.995$ to $77.032$ across the five folds, while PICP ranges from $0.947$ to $0.998$. Historical lag features remain available for held-out sites, so these folds measure warm-start site behavior rather than cold-start generalization. The variation characterizes the tested application regime; it is not a universal estimate of deployment performance.

\begin{table}[t]
\centering
\caption{Complementary stored warm-start site-held-out behavior of Bayesian TSK on BDG2. Each fold holds out sites while retaining explicitly available historical lag features; this table is separate from the Route C common-unit primary estimand and is not a cold-start test.}
\label{tab:bdg2-site}
\small
\begin{tabular}{r l r r r}
\toprule
Fold & Held-out sites & $n$ & RMSE $\downarrow$ & PICP$_{90}$ \\
\midrule
0 & Bear, Fox, Mouse, Shrew & 10800 & 13.444 & 0.987 \\
1 & Bobcat, Gator, Panther, Wolf & 9600 & 9.995 & 0.998 \\
2 & Bull, Hog, Peacock & 9000 & 77.032 & 0.969 \\
3 & Cockatoo, Lamb, Rat & 9000 & 24.239 & 0.984 \\
4 & Eagle, Moose, Robin & 9000 & 36.700 & 0.947 \\
\bottomrule
\multicolumn{5}{p{0.96\linewidth}}{\footnotesize Source: \texttt{site\_holdout\_results.json}; MPIW$_{90}$ and proper scores remain in the machine-readable result ledger.}
\end{tabular}
\end{table}

\FloatBarrier

\subsection{Calibration}
\label{sec:calibration}

\begin{figure}[htbp]
\centering
\includegraphics[width=0.85\textwidth]{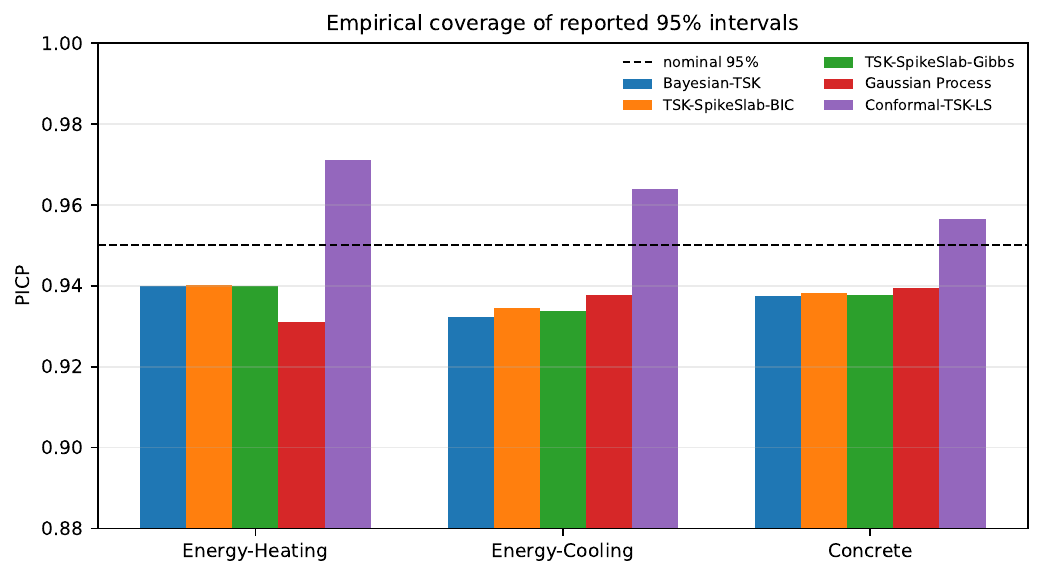}
\caption{Empirical coverage of reported 95\% intervals under the rebuilt protocol. The dashed line marks the nominal 0.95 level; split-conformal intervals use an inner calibration split and are shown as a complementary reference.}
\label{fig:cal}
\end{figure}

Fig.~\ref{fig:cal} and Table~\ref{tab:predictive} report interval coverage, width, and proper scores \cite{gneiting2007strictly}.

The conjugate Bayesian and Gibbs/BMA TSK intervals are close to, but not identical to, the nominal level.
Their PICP values are $0.940/0.932/0.938$ and $0.940/0.934/0.938$ on Heating/Cooling/Concrete, respectively.
The BIC comparator gives $0.940/0.934/0.938$; after correction, it therefore does not support the earlier undercoverage claim.
The Gibbs widths are $8.180$, $10.295$, and $29.148$, compared with Gaussian-process widths of $1.748$, $5.175$, and $20.457$.
The proper scores in Table~\ref{tab:predictive} also favor the Gaussian process on these benchmarks. The TSK intervals are consequently best interpreted as an uncertainty representation for an interpretable rule model, not as the strongest probabilistic forecast in this comparison.

\begin{table}[htbp]
\centering
\caption{Posterior-predictive metrics for methods that produce probabilistic intervals. WIS and CRPS are unavailable for the conformal baseline because it does not define a posterior-predictive distribution.}
\label{tab:predictive}
\scriptsize
\setlength{\tabcolsep}{3pt}
\begin{tabular}{llrrrr}
\toprule
Target & Method & PICP & MPIW & WIS & CRPS \\
\midrule
Energy-Heating & Bayesian-TSK & 0.940 & 8.164 & 0.938 & 1.129 \\
Energy-Heating & TSK-SpikeSlab-BIC & 0.940 & 8.151 & 0.944 & 1.138 \\
Energy-Heating & TSK-SpikeSlab-Gibbs & 0.940 & 8.180 & 0.939 & 1.132 \\
Energy-Heating & Gaussian Process & 0.931 & 1.748 & 0.210 & 0.252 \\
Energy-Cooling & Bayesian-TSK & 0.932 & 10.270 & 1.203 & 1.442 \\
Energy-Cooling & TSK-SpikeSlab-BIC & 0.934 & 10.248 & 1.205 & 1.445 \\
Energy-Cooling & TSK-SpikeSlab-Gibbs & 0.934 & 10.295 & 1.203 & 1.443 \\
Energy-Cooling & Gaussian Process & 0.938 & 5.175 & 0.597 & 0.706 \\
Concrete & Bayesian-TSK & 0.938 & 28.943 & 3.505 & 4.248 \\
Concrete & TSK-SpikeSlab-BIC & 0.938 & 28.894 & 3.505 & 4.249 \\
Concrete & TSK-SpikeSlab-Gibbs & 0.938 & 29.148 & 3.518 & 4.261 \\
Concrete & Gaussian Process & 0.939 & 20.457 & 2.447 & 2.946 \\
\bottomrule
\end{tabular}
\end{table}

\begin{table}[htbp]
\centering
\caption{Split-conformal TSK-LS results. Calibration is performed inside each outer training split, so this row is not a posterior-predictive baseline under the same protocol as Table~\ref{tab:predictive}.}
\label{tab:conformal}
\scriptsize
\begin{tabular}{lrrr}
\toprule
Target & $R^2$ & PICP & MPIW \\
\midrule
Energy-Heating & 0.956 & 0.971 & 9.433 \\
Energy-Cooling & 0.922 & 0.964 & 12.524 \\
Concrete & 0.787 & 0.957 & 32.932 \\
\bottomrule
\end{tabular}
\end{table}

Split-conformal TSK-LS gives PICP $=0.971$, $0.964$, and $0.957$ with MPIW $=9.433$, $12.524$, and $32.932$.
Its higher empirical coverage comes with wider intervals and a different calibration protocol.
On the high-dimensional probe, the rule-level Gibbs interval has PICP $0.921$ and MPIW $59.868$, compared with $0.921$ and $52.884$ for the dense conjugate model (Supplementary Table~8).
Neither comparison establishes a universal calibration advantage for one estimator.

The uncertainty results should consequently be read as a geometry of trade-offs rather than a winner-takes-all table.

On the controlled benchmarks, the Gaussian-process intervals are narrower and the proper scores favor that reference, while the conformal intervals trade additional width for coverage under their calibration split. On the measured-building split, the Bayesian interval is wider and overcovers relative to its nominal level, whereas the Ridge conformal interval is closer to its nominal coverage under its own construction. The two observations are compatible because coverage and width depend on the probability object, calibration protocol, and data regime. Reporting them separately makes it possible to identify the cost of coverage without claiming that one interval family dominates in every setting.

The high-dimensional probe supplies a second boundary condition. Rule-level and dense Bayesian intervals have the same displayed PICP but different widths, so equal empirical coverage does not imply equivalent uncertainty behavior. The coefficient-level sensitivity likewise approaches the dense reference only at diffuse slab settings without exceeding it. These results make sparsity diagnostics useful even when they do not yield a positive accuracy result: they show whether a compact rule structure changes the uncertainty representation, and they expose when that change is not supported by the tested data.

\FloatBarrier

\subsection{The Sparsity Boundary}
\label{sec:sparse}

The sparsity analysis asks whether selection changes the result.
In the 30-split main experiment, the rule-level Gibbs model retains a mean active-rule ratio of $0.713$, $0.700$, and $0.800$ on Heating, Cooling, and Concrete.
Selection is therefore substantial, but it does not improve prediction: its $R^2$ is slightly below TSK-LS on all three targets.
The result is not that the posterior is always fully active; the selected rule configurations simply do not yield a stable accuracy gain in these data.

\begin{figure}[tbhp]
\centering
\includegraphics[width=1.0\textwidth]{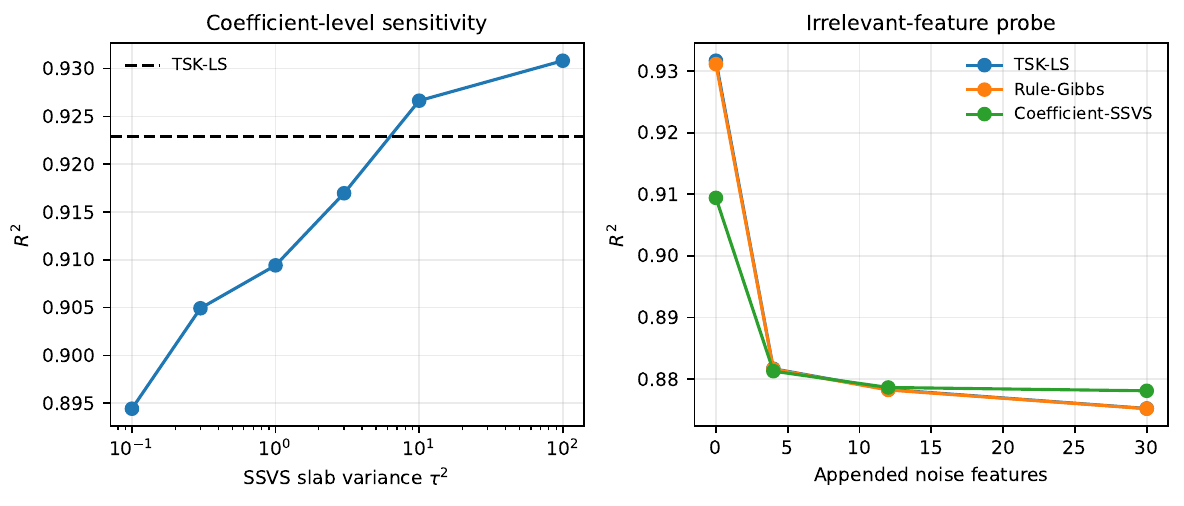}
\caption{Sparsity sensitivity under the rebuilt protocol on Energy-Cooling. The left panel varies the coefficient-level slab variance; the right panel appends standard-normal noise features. Values are regenerated from the recorded 10-split ablations and show no stable improvement over the corresponding dense reference.}
\label{fig:sparse}
\end{figure}

The coefficient-level sensitivity appears in Fig.~\ref{fig:sparse} and Supplementary Table~6.
On the 10-split Energy-Cooling ablation, $R^2$ rises from $0.894$ at $\tau^2=0.1$ to $0.931$ at $\tau^2=100$.
The dense 10-split reference under the same protocol is $R^2=0.932$ (the zero-noise row in Supplementary Table~7), so the diffuse slab approaches but does not exceed that reference.

Appending standard-normal noise features changes dense TSK-LS from $R^2=0.932$ at zero appended features to $0.875$ at 30; the sparse variants remain close, without consistent improvement across noise levels.
The appropriate conclusion is therefore protocol-bounded, not a universal sparsity boundary.

\FloatBarrier

%============================================================
\section{Discussion}
\label{sec:discussion}

\subsection{What the Rebuild Changes}

The rebuilt evidence does not support a single-cause explanation for the earlier negative results.
With corrected membership, prior, and numerical-stability choices, all four ablation configurations remain close on Energy-Cooling: BIC-BMA gives $R^2=0.937$, BIC-PIP thresholding with Laplace prediction gives $0.934$, Gibbs-PIP thresholding gives $0.933$, and the full BIC comparator gives $0.931$ (Supplementary Table~5).
The small ordering difference is compatible with a modest benefit from averaging in this ablation, but it is not evidence of universal BMA dominance and cannot attribute the earlier collapse to thresholding or covariance alone.
The implementation audit is therefore part of the scientific comparison: membership persistence, prior scaling, and numerical linear algebra determine the object being compared.

\subsection{A Correctness-First Evaluation Principle}
\label{sec:principle}

The central lesson of the rebuild is that correctness and performance cannot be evaluated on separate tracks when the model representation carries the uncertainty calculation. In a weighted TSK regression, the antecedent partition is part of the design matrix, and the design matrix is part of the posterior and predictive distribution. A silent change in the partition can therefore produce a numerically plausible prediction from a scientifically different fitted object. The failure-mode reproduction makes this dependency visible, while the corrected pipeline provides the stable object against which the Bayesian variants can be compared.

Three consequences follow from this dependency. First, representation invariants should be stored and tested as model state. The frozen membership parameters, no-intercept construction, scaled prior, stable solves, and log-domain firing strengths are not interchangeable implementation preferences; they determine which model is being evaluated. Second, uncertainty claims should identify the layer at which they are made. Posterior-predictive quantiles describe a Bayesian model and its finite draws, while conformal intervals describe a calibration procedure. PICP, MPIW, WIS, CRPS, and point error answer different questions and should remain visible together. Third, external validity should be attached to the sampling unit. Repeated tabular splits, a chronological building sample, and warm-start site folds provide different forms of evidence and should not be collapsed into a single score.

This principle gives the negative results a constructive role. The selected rule configurations are active in the controlled benchmark, yet their point accuracy remains close to or below dense references. The application split shows that persistence can remain the strongest point reference while a Bayesian interval is wider and overcovers. Rather than weakening the evaluation, these findings identify where a sparse uncertainty workflow changes the representation without delivering a measured predictive benefit. The resulting knowledge is a reproducible boundary condition: selection can be computationally and diagnostically meaningful without being a universal accuracy or calibration mechanism.

The principle is also portable as an evaluation procedure. An applied soft-computing study using rule-based uncertainty can begin by freezing the representation, then validate the algebra on a small controlled case, compare posterior and non-posterior uncertainty objects under their own protocols, report behavior at the correct temporal or group unit, and retain failure cases beside successful runs. The present code and supplementary tables instantiate that sequence for the stated TSK model and datasets. They do not establish that the sequence is sufficient for every fuzzy system, but they make each transfer step explicit enough to be tested rather than assumed.

\subsection{Significance for Applied Soft Computing}
\label{sec:asocsignificance}

The significance of this study lies in connecting soft-computing representation choices to the evidence needed for an applied uncertainty claim. Fuzzy rules provide a compact, interpretable structure, but interpretability does not by itself validate the uncertainty attached to that structure. The results show why an applied study should report both the rule representation and the behavior of its predictive distribution: the same model family can remain interpretable, match dense TSK point accuracy, produce wider intervals, and still fail to improve the measured point reference.

This framing distinguishes three outcomes that are often merged in applied comparisons. An implementation may be correct without being more accurate. A predictive interval may have high empirical coverage without being efficient in width. A selected rule set may provide a compact or inspectable configuration without improving the target metric. The current evidence exhibits all three separations across controlled and measured settings. Making them explicit gives readers a more useful basis for deciding whether the workflow fits a new application than a single average rank would provide.

For practitioners, the practical implication is a reporting protocol rather than a deployment recommendation. A new application should preserve training-time antecedent parameters, identify whether site or temporal features imply warm-start information, state which uncertainty object is being calibrated, and show the width and point-error cost alongside coverage. The Route C BDG2 results demonstrate why this matters: a common-unit point comparison still shows no Bayesian TSK advantage, the posterior interval is wider and overcovers relative to the conformal reference, and conditional effects vary across sites and load regimes. These observations support careful reuse of the evaluation logic while leaving deployment utility, cold-start transfer, and intervention effects to separately designed studies.

\subsection{Implications for Sparse TSK Research}

The sparse results depend on data geometry and prior scale.
Rule-level selection is active in the low-dimensional benchmark, but it is not a reliable source of predictive improvement.
In the real high-dimensional probe, dense TSK-LS and the dense Bayesian model obtain $R^2=0.820$ and $0.822$, whereas rule-level Gibbs obtains $0.764$.
Coefficient-level SSVS gives $0.813$ at $\tau^2=1$ and $0.818$ at $\tau^2=10$; the complete grid reaches $0.819$ at its most diffuse setting, below the corresponding dense reference of $0.821$.
These results motivate reporting active-set and sensitivity diagnostics with sparsity, but they do not identify a general threshold at which selection becomes beneficial.

Alternative shrinkage priors, such as the continuous spike-and-slab LASSO \cite{rockova2018spikeslab} and the horseshoe \cite{carvalho2010horseshoe}, may behave differently under collinearity in a TSK weighted design matrix.

\subsection{What the Real-Building Benchmark Adds}

The BDG2 evaluation expands the application evidence without changing the method claim. It adds measured hourly electricity, multiple sites, chronological extrapolation, missing windows, a common-unit comparison, and a complementary warm-start site-held-out analysis. In this setting, persistence remains the strongest point-prediction reference under the common Route C rows, while the Bayesian TSK posterior provides a wider uncertainty representation with overcoverage on the main test split. Real building meters therefore expose conditional behavior, but these results do not establish energy savings, control utility, or a universally calibrated predictor.

The application result is useful precisely because the protocol records the information available to the model. Historical lags remain present for held-out sites, so the site analysis measures behavior under a warm-start condition rather than an unseen-building cold start. The bounded sample also separates data eligibility from test performance: buildings are selected by the recorded target-coverage rule and fixed hash order, not by a favorable outcome. This makes the negative point result interpretable as a property of the stated sample and feature policy, rather than as a claim about every building-meter deployment.

\subsection{Limitations}

The conclusions are bounded in several ways.
The controlled evaluation contains two UCI datasets, three targets, a fixed FCM antecedent structure, and five rules; the Superconductivity experiment is a 3,000-row probe rather than a broad high-dimensional study.
Raw coefficient diagnostics show \(\hat R_{\max}>1.03\) on representative Gibbs runs, so finite-chain uncertainty and possible mixing problems remain relevant even though predictive-quantity diagnostics are smaller.
The conformal comparator uses an inner calibration split and is not protocol-equivalent to the posterior-predictive intervals.
Sparsity is applied to consequents only; joint antecedent--consequent selection \cite{gu2018bayesian} and alternative shrinkage priors remain open.

Finally, BDG2 is a bounded sampled evaluation with an explicit warm-start policy and mixed upstream sources. It does not measure operational intervention, energy savings, or cold-start transfer, and its building-level metadata should not be treated as a public release of raw records.

%============================================================
\section{Conclusion}
\label{sec:conclusion}

We rebuilt a Bayesian TSK regression pipeline with explicit membership persistence, scaled priors, stable linear algebra, block-Gibbs sampling, and posterior-predictive model averaging.
Across three UCI targets, Gibbs/BMA remained close to the dense and conjugate TSK references in point accuracy, with empirical 95\% coverage close to but not exactly at the nominal level.
The bounded BDG2 evaluation extends the evidence to measured building meters through chronological and warm-start site-held-out tests: persistence remains the strongest simple point reference, while Bayesian TSK intervals are wider and overcover on the main sampled test split.
The Gaussian process is stronger on point accuracy and proper probabilistic scores in the controlled benchmarks, while split conformal produces wider intervals with higher empirical coverage.
The rebuilt BIC comparator does not reproduce the earlier negative-$R^2$ or severe-undercoverage behavior.
The resulting conclusion is implementation- and protocol-bounded: rule-level spike-and-slab model averaging is reproducible and diagnostically useful, but its predictive advantage over dense TSK in the controlled evaluation and over persistence in the common-unit building-meter evaluation is not established.

\section*{CRediT authorship contribution statement}

\textbf{Zhiren Xiao}: Conceptualization, Methodology, Software, Validation, Formal analysis, Investigation, Data curation, Writing -- original draft, Writing -- review \& editing, Visualization.

\section*{Acknowledgments}

The author declares no specific funding for this research.

\section*{Data Availability}

Datasets are publicly available from the UCI Machine Learning Repository: Energy Efficiency (dataset 242; \href{https://archive.ics.uci.edu/dataset/242/energy+efficiency}{UCI dataset page}) \cite{tsanas2012accurate} and Concrete Compressive Strength (dataset 165; \href{https://archive.ics.uci.edu/dataset/165/concrete}{UCI dataset page}) \cite{yeh1998modeling}. The measured building benchmark is the Building Data Genome Project 2 v1.0 data object \cite{miller2020bdg2,miller2020bdg2zenodo}; the local analysis uses the cleaned electricity file and does not redistribute the raw archive.
The raw BDG2 archive is not redistributed here because it is third-party dataset material; readers should obtain it from the cited Zenodo v1.0 record under its stated terms. The derived split-level results and corrected code have not been publicly deposited in this run, and any future release requires an author-approved repository and licence decision.
The earlier public code archive is available at \texttt{https://github.com/Jackxiaozhiren/tsk-spikeslab} and \href{https://doi.org/10.5281/zenodo.21929319}{10.5281/zenodo.21929319}.
The current rebuilt branch and its generated results, including the BDG2 Route C artifacts, are not publicly archived with this preprint. The earlier public code archive is not a complete computational reproduction of all results reported here. The current Route C provenance, configuration, and result manifests remain in the local working artifacts; the earlier Route B result is historical bounded evidence.

\section*{Code Availability}

The corrected source code, tests, result manifests, and table/figure generator for the rebuilt branch are maintained in local working artifacts. No new public code release accompanies this preprint. The earlier repository and archive linked above do not include the complete BDG2 Route C workflow.

\section*{Declaration of Competing Interests}

The author declares no competing interests.

\section*{Declaration of Generative AI and AI-assisted technologies in the writing process}

During the preparation of this work, the author used Claude (Anthropic) for code development assistance and language editing. OpenAI Codex (OpenAI) assisted subsequent manuscript and source-package checks and preprint preparation. The author is responsible for the scientific claims, references, results, and final manuscript.

\bibliographystyle{elsarticle-num}
\bibliography{references}

\end{document}